\documentclass{article}

\usepackage{times}
\usepackage{graphicx} % more modern
\usepackage{subcaption}
\usepackage{epstopdf}
\usepackage{natbib}

\usepackage{algorithm}
\usepackage{algorithmic}
\usepackage{amsfonts}

\usepackage{amsmath}
\usepackage{hyperref}

\newcommand{\given}{\,|\,}

\renewcommand{\vec}[1]{\ensuremath{\mathbf{#1}}}
\newcommand{\R}{\ensuremath{\mathbb{R}}}
\newcommand{\mat}[1]{\ensuremath{\mathbf{#1}}}

\graphicspath{{figures/}}
\usepackage[accepted]{icml2015}

\icmltitlerunning{Active Learning by Statistical Leverage Sampling}

\begin{document} 

\twocolumn[
\icmltitle{ALEVS: Active Learning by Statistical Leverage Sampling}

% unless you've provided the [accepted] option to the icml2015
\icmlauthor{Cem Orhan}{cem.orhan@bilkent.edu.tr}
\icmladdress{Department of Computer Engineering, Bilkent University, Ankara, Turkey, 06800}
\icmlauthor{Oznur Tastan}{oznur.tastan@cs.bilkent.edu.tr}
\icmladdress{Department of Computer Engineering, Bilkent University, Ankara, Turkey, 06800}

\icmlkeywords{Active learning, Statistical leverage, Classification}

\vskip 0.3in
]

\begin{abstract} 
Active learning aims to obtain a classifier of high accuracy by using fewer label requests in comparison to passive learning by selecting effective queries. Many active learning methods have been developed in the past two decades, which sample queries based on informativeness or representativeness of unlabeled data points. In this work, we explore a novel querying criterion based on statistical leverage scores. The statistical leverage scores of a row in a matrix are the squared row-norms of the matrix containing
its (top) left singular vectors and is a measure of influence of the row on the matrix. Leverage scores have been used for detecting high influential points in regression diagnostics \cite{Chatterjee1986} and have been recently shown to be useful for data analysis \cite{dmm2008cur} and randomized low-rank matrix approximation algorithms \cite{gittens2013}. We explore how sampling data instances with high statistical leverage scores perform in active learning. Our empirical comparison on several binary classification datasets indicate that querying high leverage points is an effective strategy.
\end{abstract} 
\section{Introduction}
\label{intro}

A passive supervised learning algorithm for classification induces a model with the available set of labeled instances. However, in many modern machine learning applications, in addition to this limited set of labeled instances, there is a large pool of unlabeled instances. For cases where the cost of labeling data is high relative to that of collecting the unlabeled data, active learning strategies have been shown to be useful. In a classical active learning framework for supervised classification \cite{Cohn1994, Settles2009}, the learner can interact with an oracle (i.e. human annotator) that provides labels when queried. Typically, an active learner begins with a small set of labeled instances, selects one or a batch of examples from a pool of unlabeled data and queries the labels for these selected examples. Once the oracle provides the new labels, these examples are augmented to the training set; the active learner is retrained, and this process is repeated until a halting criterion (i.e. desired accuracy) is satisfied. Through selectively deciding which examples to label, the active learner aims to obtain a classifier of high accuracy by using fewer label requests and thereby reducing the total labeling cost. Different strategies \cite{Settles2009} of querying examples have been suggested. In this work, we explore a novel direction for querying that is based on statistical leverage scores.

The statistical leverage has found extensive applications in diagnostic regression analysis \cite{Chatterjee1986, Hoaglin1978}. Statistical leverage scores have been recently shown to be useful for data analysis such as CUR decomposition  and randomized low-rank matrix approximation algorithms. In CUR decomposition, the matrix is approximated with a product $\mat{C}\mat{U}\mat{R}$, where $\mat{C}$ and $\mat{R}$ are respectively small subsets of the columns and rows of the matrix $\mat{U}$ is computed from $\mat{C}$ and $\mat{R}$ \cite{dkm2006}. \cite{dmm2008cur} introduced a method where the matrix columns are sampled randomly with probability proportional to their leverage scores. Similarly, Nystr\"om extensions are sampling based randomized low-rank approximations to positive-semidefinite matrices. Gittens et al. analyzed different Nystr\"om sampling strategies for SPSD matrices and showed that samplings based on leverage scores are quite effective \cite{gittens2013}. 

In the aforementioned work, leverage scores were used for approximation purposes. The intuition in these methods is that leverage score sampling ensures important columns (or rows) are included in the approximation. In this study we instead exploit leverage scores to find examples with important feature vectors in the data and query the instances with high statistical leverage scores. Our proposed method, \textbf{A}ctive Learning by Statistical \textbf{Lev}erage \textbf{S}ampling (ALEVS), exhibits good empirical performance on different benchmark datasets. The rest of the paper is organized as follows: in section 2, we describe the problem set up and our approach ALEVS; in section 3, the experiments are described in detail; in section 4 we discuss the empirical performance of ALEVS on different datasets; in section 5 results are elaborated on and the conclusions are stated.

\section{Problem Set Up and Approach}
\subsection{Problem Set Up}

We denote $ \mathcal{D} = \{ (\vec{x}_1, y_1), (\vec{x_2}, y_2), \ldots, \vec{x}_n, y_n ) \}$ the training data set that contains $n$ instances, where each instance $\vec{x}_i = [x_{i1}, x_{i2}, \ldots, x_{x_id}]$ is a vector of $d$ dimension and $y_i \in \{-1, 1 \}$ is the class label of $\vec{x}_i$. The initial dataset comprises a small set of labeled examples, and a large pool of unlabeled examples. At each iteration $t$ of active learning, a perfect oracle $\mathcal{O}$ is queried with an unlabeled example $\vec{x}_q$ and the oracle returns the label $y_q$ with uniform cost across examples. We denote the labeled set of training examples at iteration $t$ with $\mathcal{D}_l^t$ and the set of unlabeled examples with $\mathcal{D}_u^t$. Our aim is to attain a good accuracy classifier $h^*$ with minimal number of queried examples.

\subsection{ALEVS: Sampling Based on Statistical Leverage Scores}
\label{sec:alevs}

At an iteration $t$, the classifier, $h_t$ is trained only with the labeled training examples $\mathcal{D}_l^t$ and the data is divided into two portions based on class memberships. Two feature matrices are formed. $\mat{X}_+^t$ is a $m \times d$ feature matrix, where the rows are the feature vectors of examples with positive class membership at iteration $t$. These examples are those that are positively labeled in $\mathcal{D}_l^t$ and those that are  in $\mathcal{D}_u^t$ but have predicted positive labels according to $h_t$. $\mat{X}_-^t$ is similarly constructed from negatively predicted and labeled examples.

%matrix indexing için matlab-like birşey yazdım, onun düzeltilmesi gerekiyor, nasıl olması gerektiğini bulamadım
\begin{algorithm}[!ht]
	\caption{\textsc{alevs}: Active Learning with Leverage Score Sampling}
	\label{alg:ALEVSKernel}
	\begin{algorithmic}
		\STATE {\bf Input:}  $D$ a training dataset of $n$ instances; Labeling oracle $\mathcal{O};$ low-rank parameter $k$; kernel parameters if any
		\STATE {\bf Output:}  Classifier $h^*$
		\STATE {\bf Initialize:}  \\
		 \STATE $\mathcal{D}_l^{0}$ \quad \quad \quad \quad  \quad  \quad \quad \% initial set of labeled instances
		 \STATE$\mathcal{D}_u^{0} \gets \mathcal{D} \setminus \mathcal{D}_l^{0}$ \quad\quad\quad \% the pool of unlabeled instances
		 \REPEAT
		 \STATE ------------------ {\bf Classification} ---------------------------
		\STATE Train classifier $h_t $  with training data $D_l^t$
		\STATE Get predicted class labels $\vec{\hat{y}_u^t}$  by applying $h_t$ on  $\mathcal{D}_u^t$
		 \STATE ------------------ {\bf Sampling} ---------------------------------
		 \STATE  Based on $\vec{\hat{y}}_u^t$ and $\vec{y}_l^t$, construct $\mat{X}_{+}^t$ and  $\mat{X}_{-}^t$ 
		\STATE Compute  kernel matrix $\mat{K}_{+}^t$ on $\mat{X}_+^t$ 
		\STATE  Compute kernel matrix   $\mat{K}_-^t$  on $\mat{X}_-^t$  
		\STATE Compute leverage scores on  $\mat{K}_+^t$  using Eq. \ref{eq:levscore}
	          \STATE Compute leverage scores  on $\mat{K}_-^t$  using Eq. \ref{eq:levscore}
	          \STATE Get $x_q$ with the highest leverage score in $\mathcal{D}_u^t$
		\STATE Query $\mathcal{O}$ its label $y_q$
		 \STATE ------------------ {\bf Update} ----------------------------------- 
		\STATE $\mathcal{D}_l^{t+1} \gets  \mathcal{D}_l^t \cup (\vec{x_q}, y_q)$ 
		\STATE $\mathcal{D}_u^{t+1} \gets \mathcal{D}_u^t \setminus \vec{x_q}$ 
		\STATE $t \gets t+1$
		\UNTIL stopping criterion
		\STATE $h^* \gets h_t$
		\STATE Return $h^*$
	\end{algorithmic}
\end{algorithm}

After the prediction of the labels of unlabeled data, ALEVS computes a kernel matrix over $\mat{X}_+^t$ and $\mat{X}_-^t$ separately. In our experiments we employed linear kernel and Gausian Radial Basis (RBF) kernel.  Over a set of data points $\vec{x}_1, \ldots, \vec{x}_n \in \R^d, $ the linear kernel 
matrix $\mat{K}$ corresponding to those points is given by

\begin{equation}
\label{eq:linkernel}
K_{ij} = \langle \vec{x}_i, \vec{x}_j \rangle.
\end{equation}
RBF kernel matrix $\mat{K}$ corresponding to these same points
is given by
\begin{equation}
\label{eq:rbfkernel}
K_{ij} = \exp\bigg(\frac{-\left\|{\vec{x}_i - \vec{x}_j}\right\|_2^2}{2\sigma^2} \bigg).
\end{equation}
In the above equation $\sigma$ is a nonnegative real number that determines the scale of the kernel. The choice of $\sigma$ is discussed in the experimental section.

As described in \cite{gittens2013}, the leverage scores of a SPSD  kernel matrix $\mat{K}\in \R^{nxn}$ can be calculated as follows.  $\mat{K}=\mat{U}\mat{\Sigma} \mat{U}^T$ is the eigen decomposition of $\mat{K}$. We can partition $\mat{U}$ as 
\begin{equation}
\mat{U} = \bigg (\mat{U}_1 \quad \mat{U}_2 \bigg), 
\end{equation}
  where $\mat{U}_1$ comprises $k$ orthonormal columns spanning  the top $k$-dimensional eigenspace of $\mat{K}$.   
 The leverage score of the $j$th column
of $\mat{K}$ is defined as the squared Euclidean norm of the $j$th row of $\mat{U}_1:$
\begin{equation}
\label{eq:levscore}
\ell_j = \|({U}_1)_{(j)}\|_2^2.
\end{equation}

After the leverage scores are computed within each class, the example to query $\vec{x}_q$ is determined by selecting the unlabeled example with the highest leverage score:
\begin{equation}
\vec{x}_q = \arg\max_{\vec{x}_j \in \mathcal{D}_u^{t} } \ell_j 
\end{equation}

Steps of ALEVS are summarized in  Algorithm \ref{alg:ALEVSKernel}.

\newcommand{\newW}{0.235}
\begin{figure*}[!ht]
	\centering
	\begin{subfigure}[b]{\newW\textwidth}
		\includegraphics[width=\textwidth]{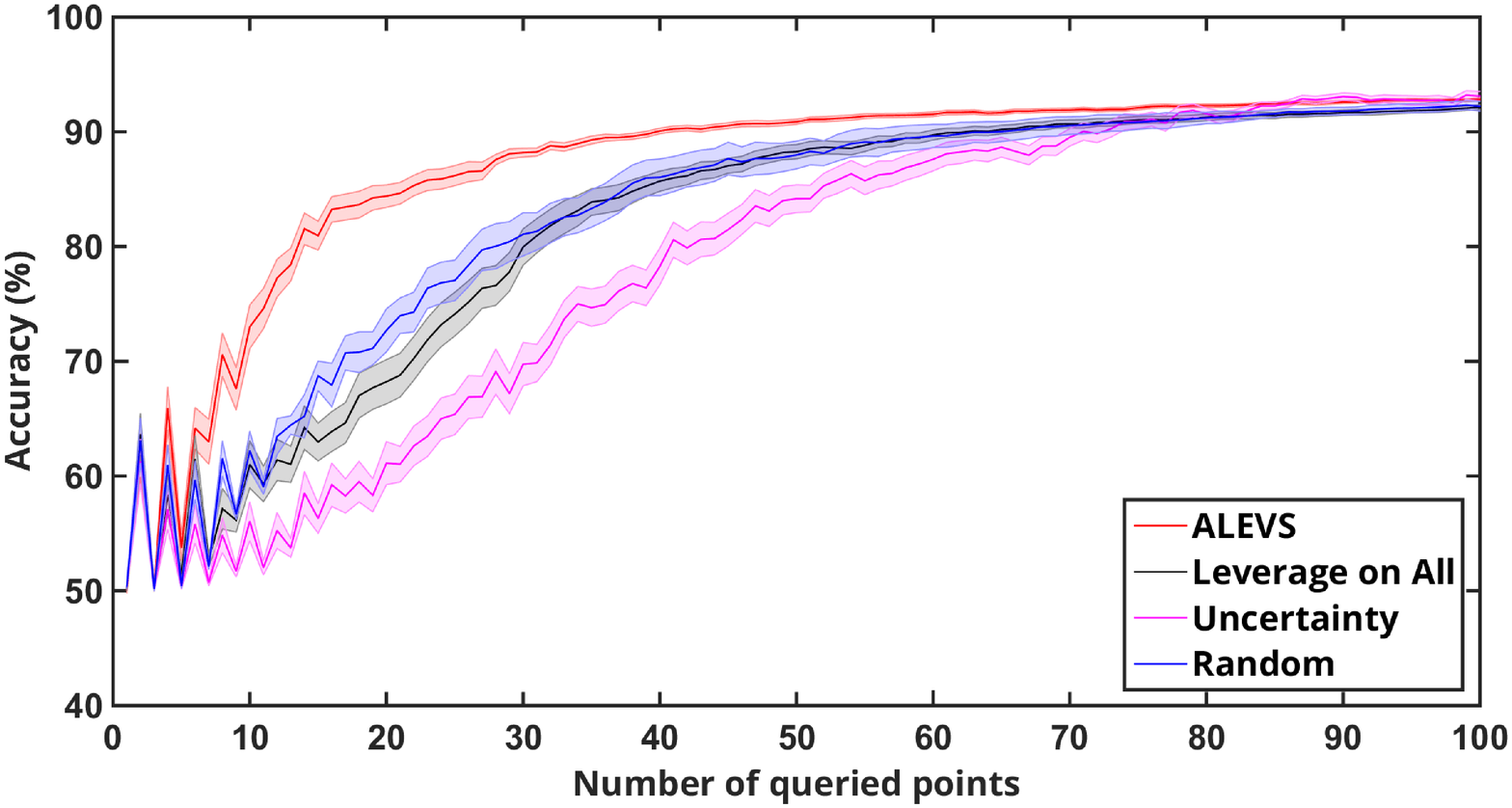}
		\caption{\textit{digit1}, $k=60$, RBF}
		\label{digit1}
	\end{subfigure}
	~
	\begin{subfigure}[b]{\newW\textwidth}
		\includegraphics[width=\textwidth]{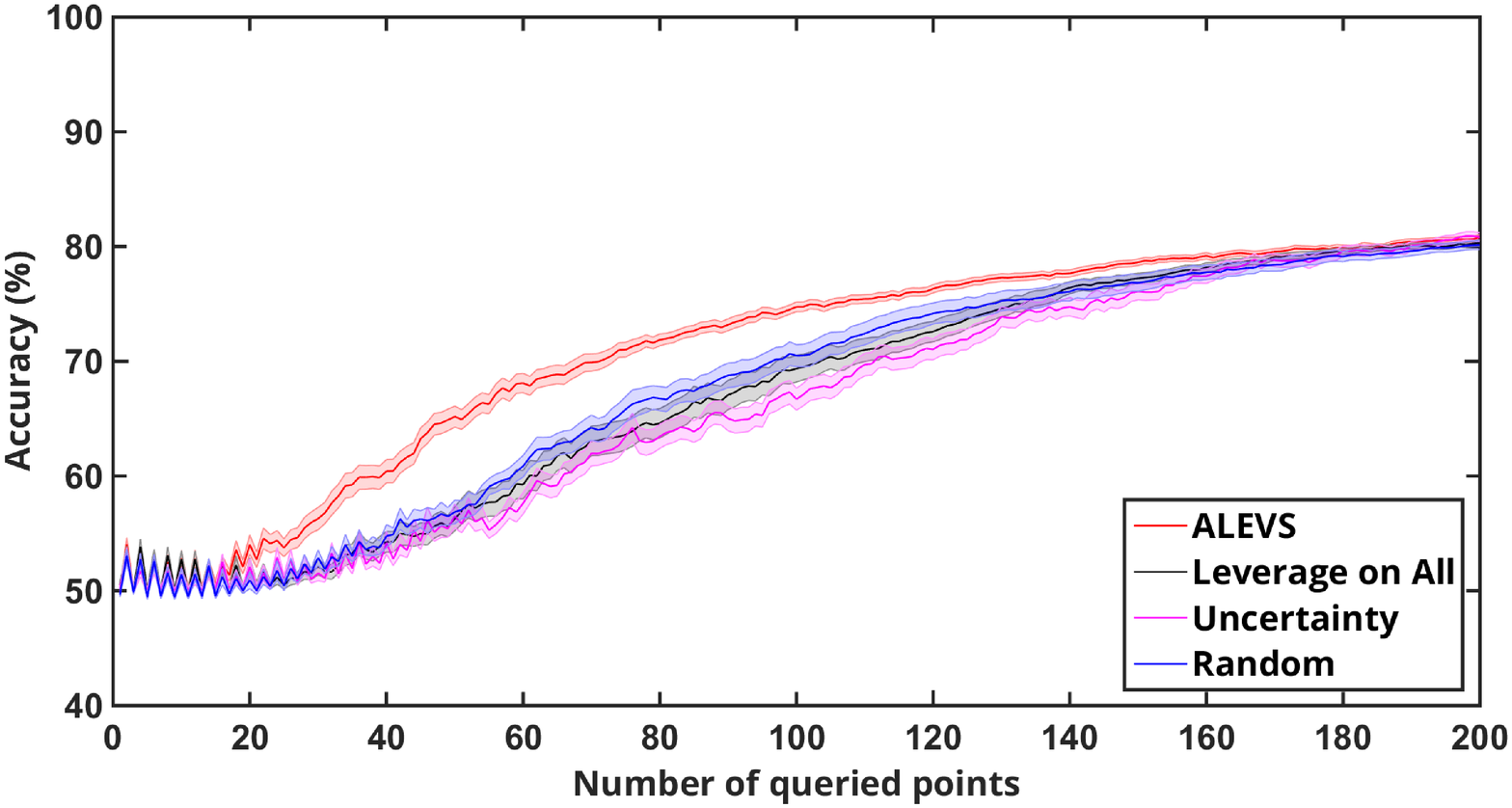}
		\caption{\textit{g241n}, $k=60$, RBF}
		\label{g241n}
	\end{subfigure}
	~
	\begin{subfigure}[b]{\newW\textwidth}
		\includegraphics[width=\textwidth]{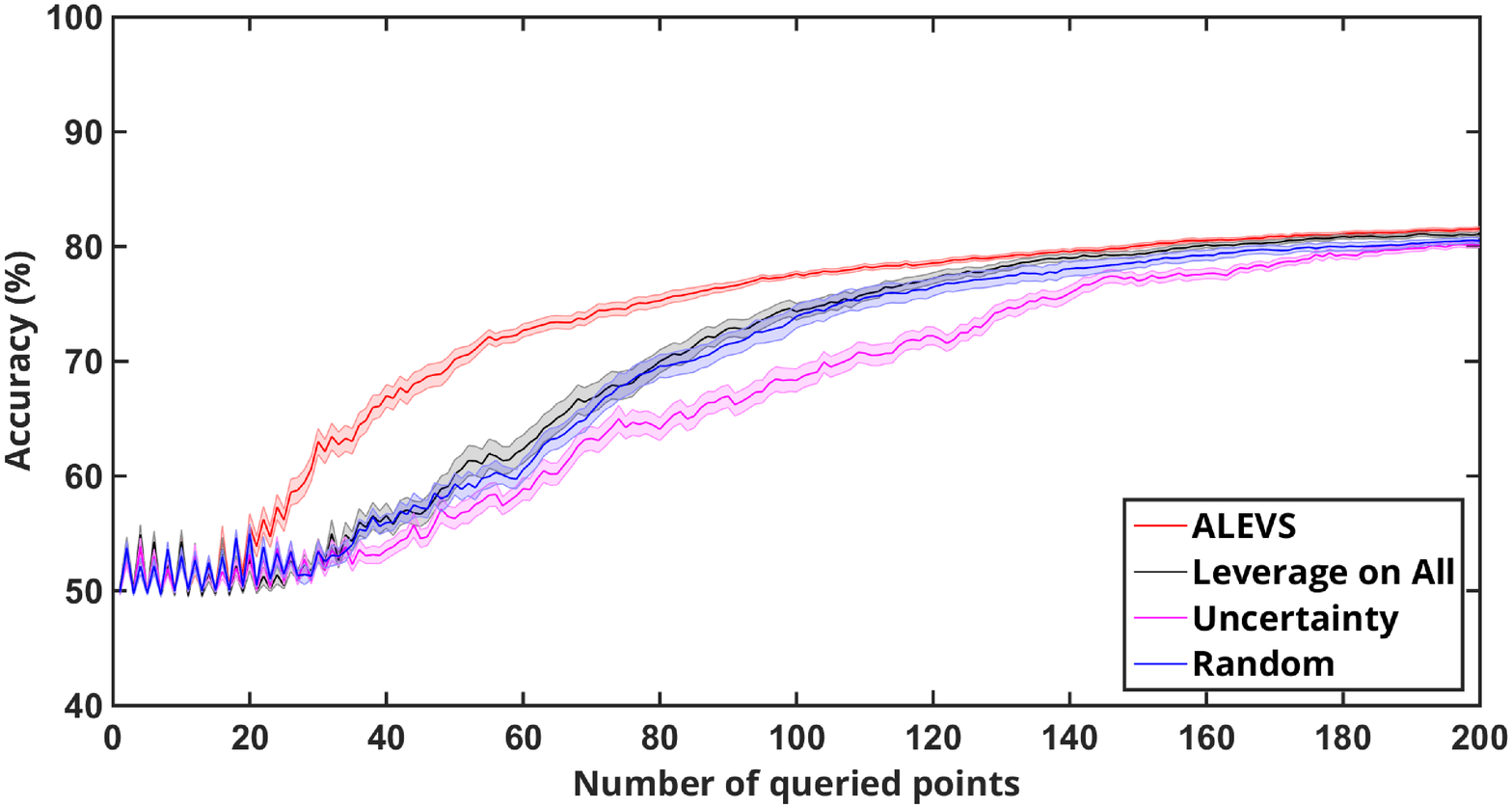}
		\caption{\textit{g241c}, $k=60$, RBF}
		\label{g241c}
	\end{subfigure}
	~
	\begin{subfigure}[b]{\newW\textwidth}
		\includegraphics[width=\textwidth]{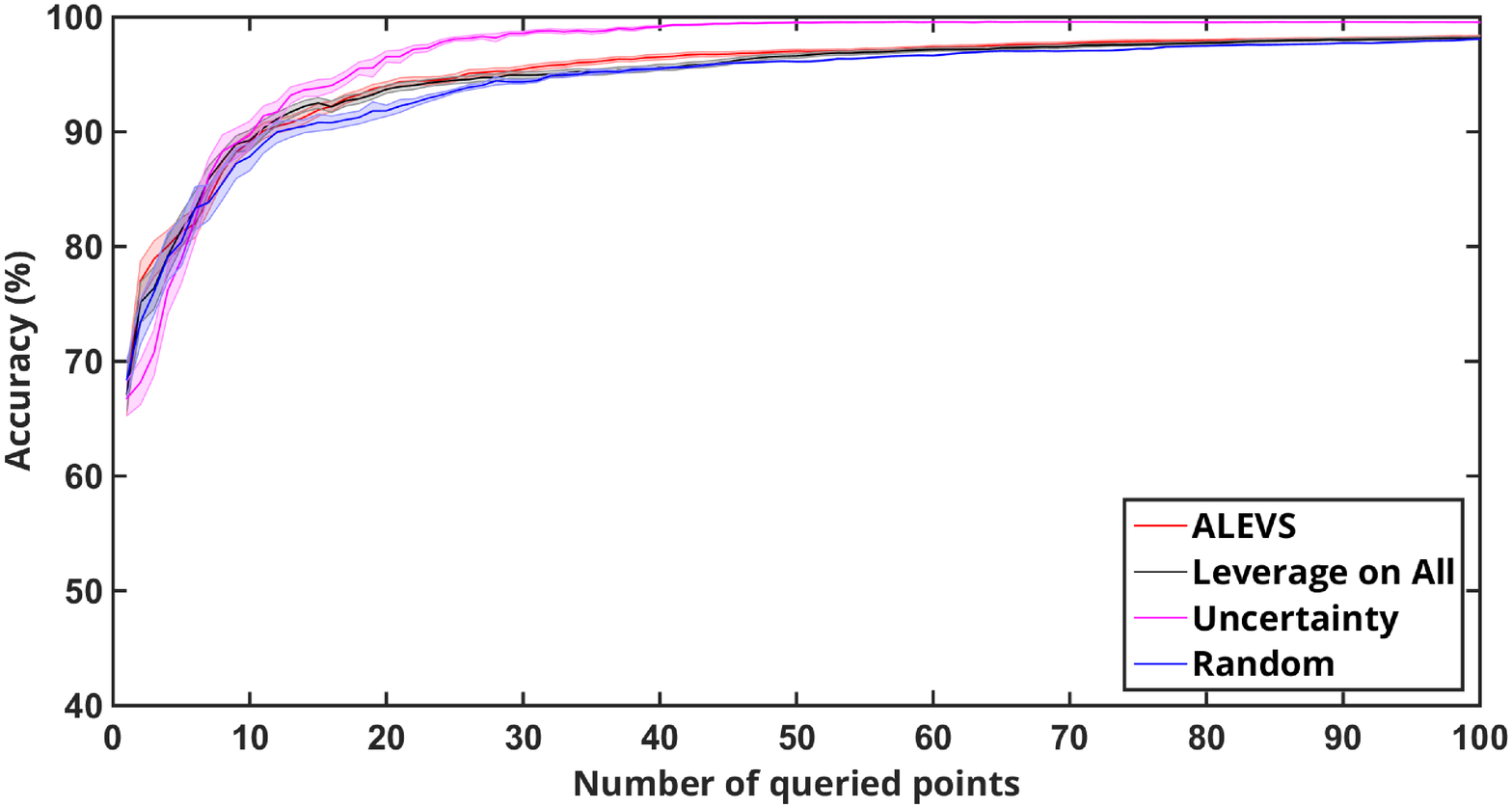}
		\caption{\textit{D vs. P}, $k=60$, RBF }
		\label{letter1}
	\end{subfigure}
	
	\begin{subfigure}[b]{\newW\textwidth}
		\includegraphics[width=\textwidth]{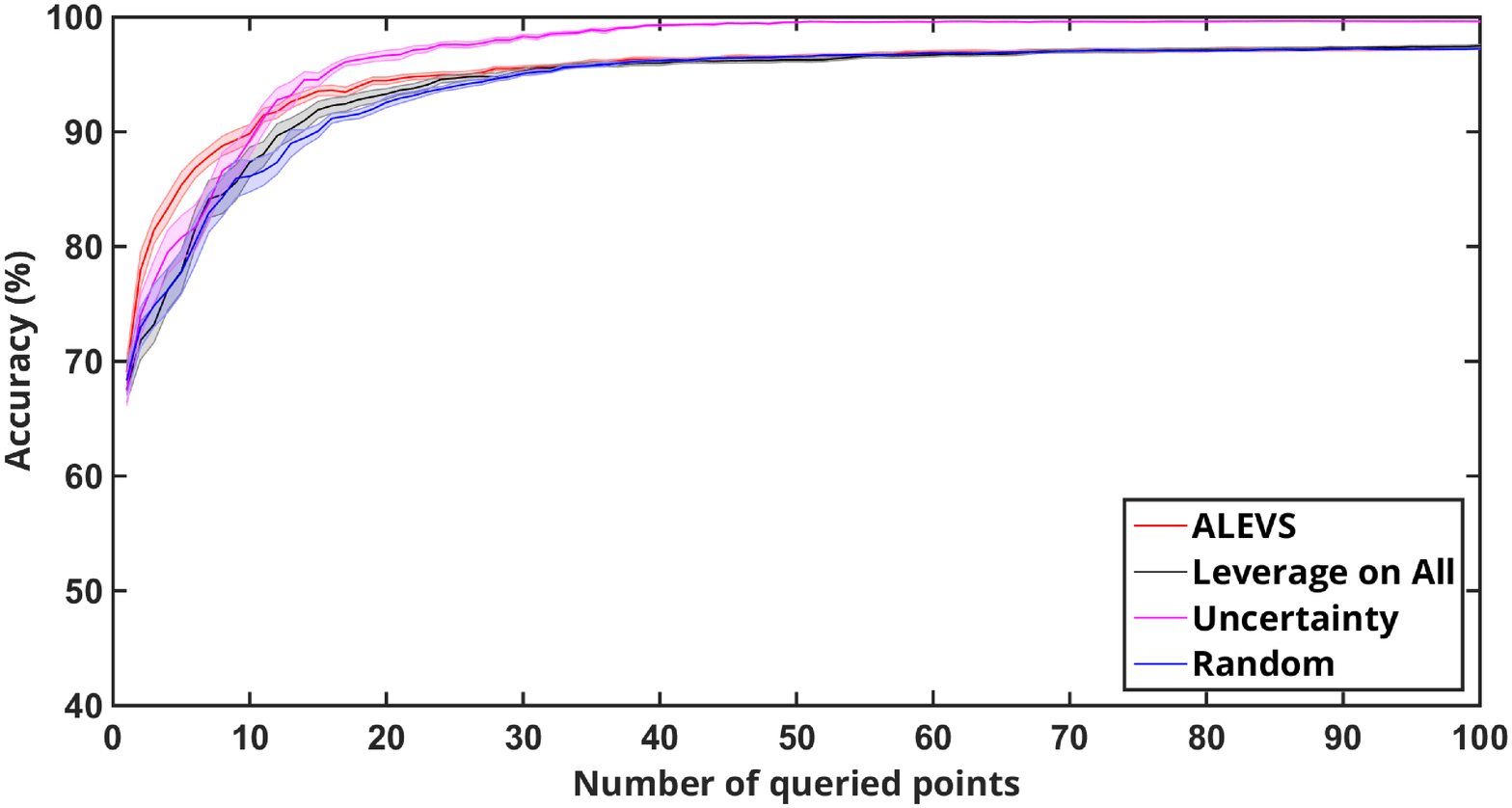}
		\caption{\textit{U vs. V}, $k=60$, RBF}
		\label{letter2}
	\end{subfigure}
	~
	\begin{subfigure}[b]{\newW\textwidth}
		\includegraphics[width=\textwidth]{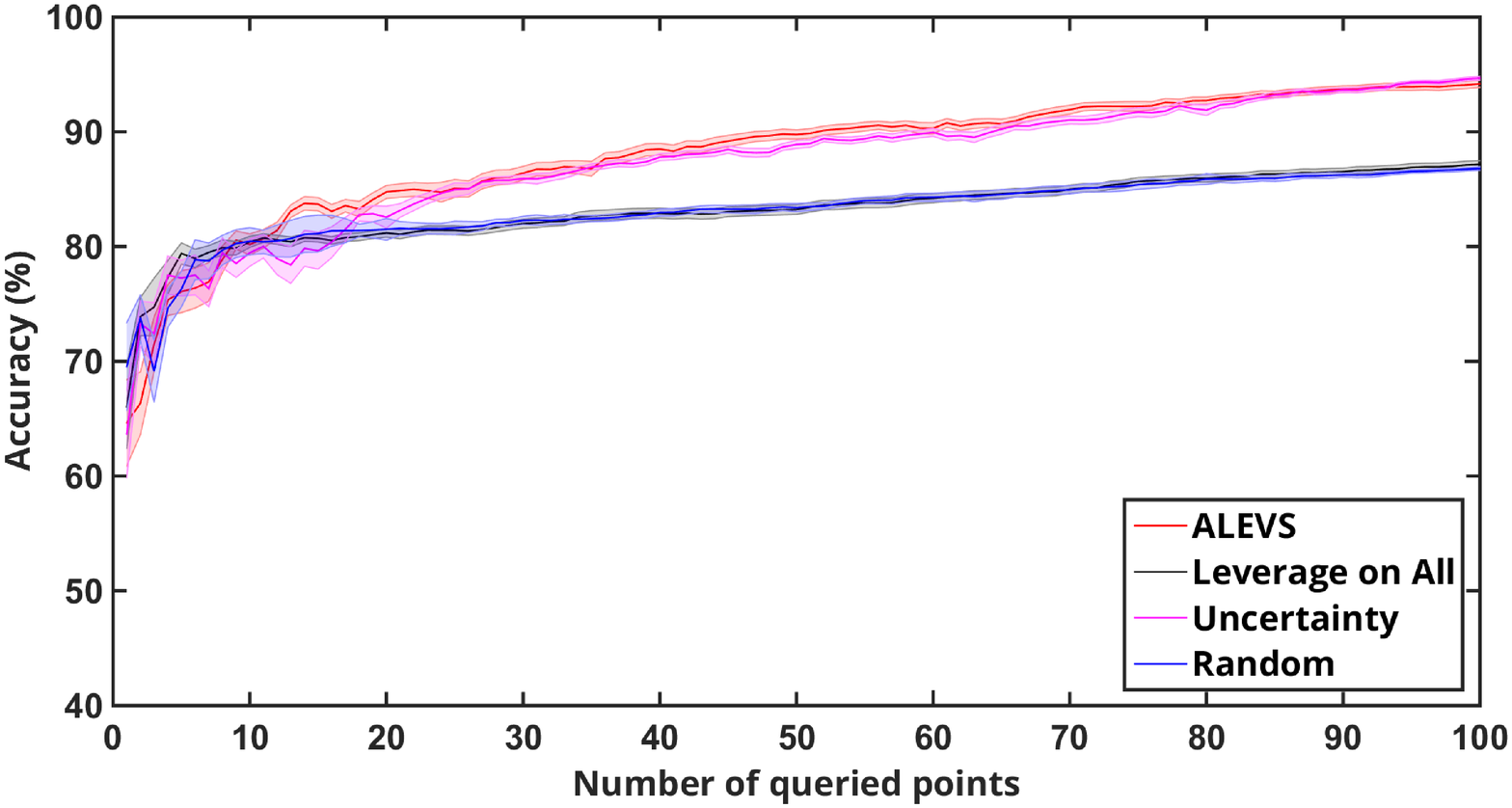}
		\caption{\textit{USPS}, $k=60$, RBF}
		\label{USPS}
	\end{subfigure}
	~
	\begin{subfigure}[b]{\newW\textwidth}
		\includegraphics[width=\textwidth]{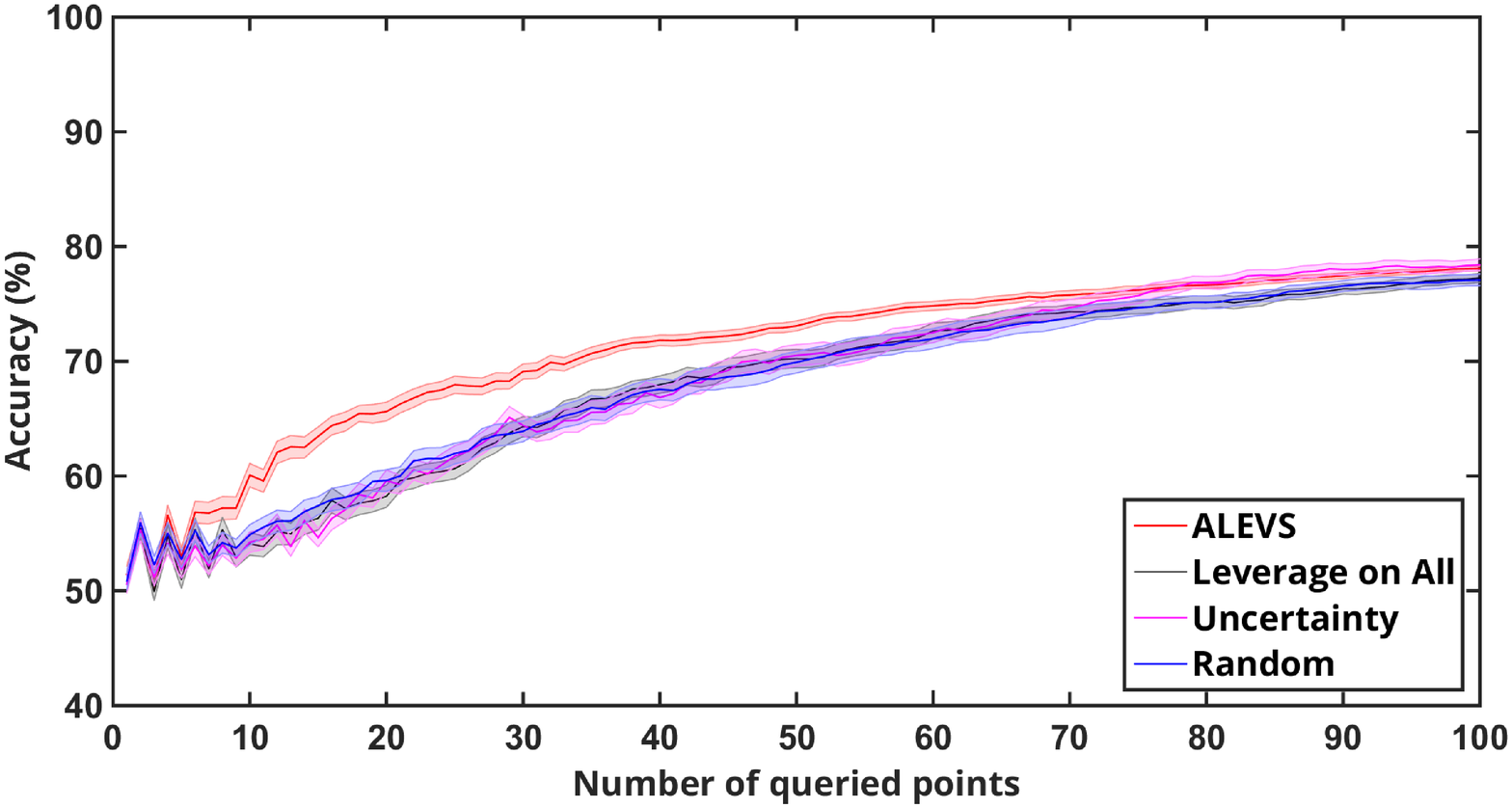}
		\caption{\textit{splice}, $k=80$, linear }
		\label{splice}
	\end{subfigure}
	~	
	\begin{subfigure}[b]{\newW\textwidth}
		\includegraphics[width=\textwidth]{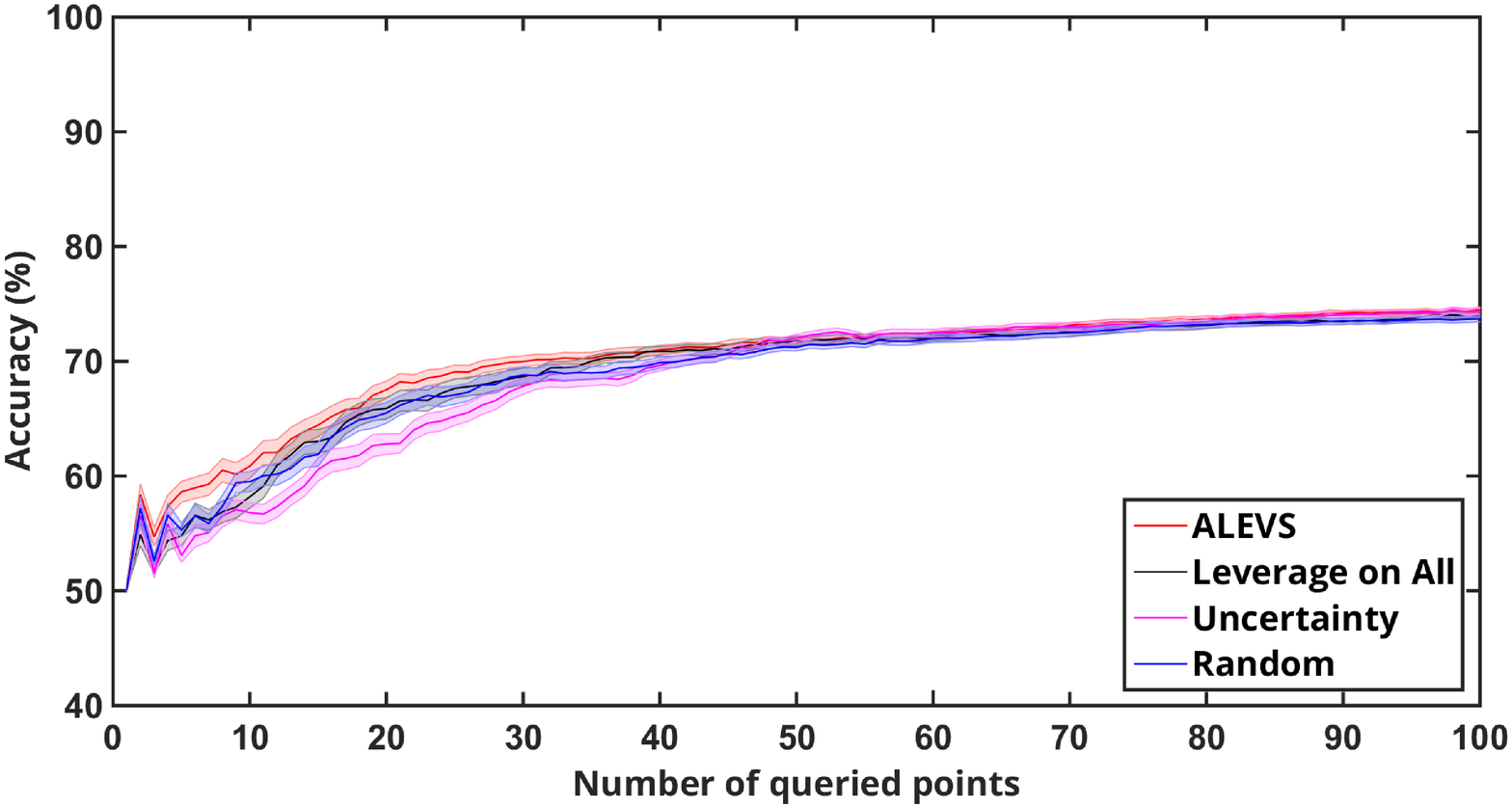}
		\caption{\textit{ringnorm}, $k=60$, RBF }
		\label{ringnorm}
	\end{subfigure}
	
	\begin{subfigure}[b]{\newW\textwidth}
		\includegraphics[width=\textwidth]{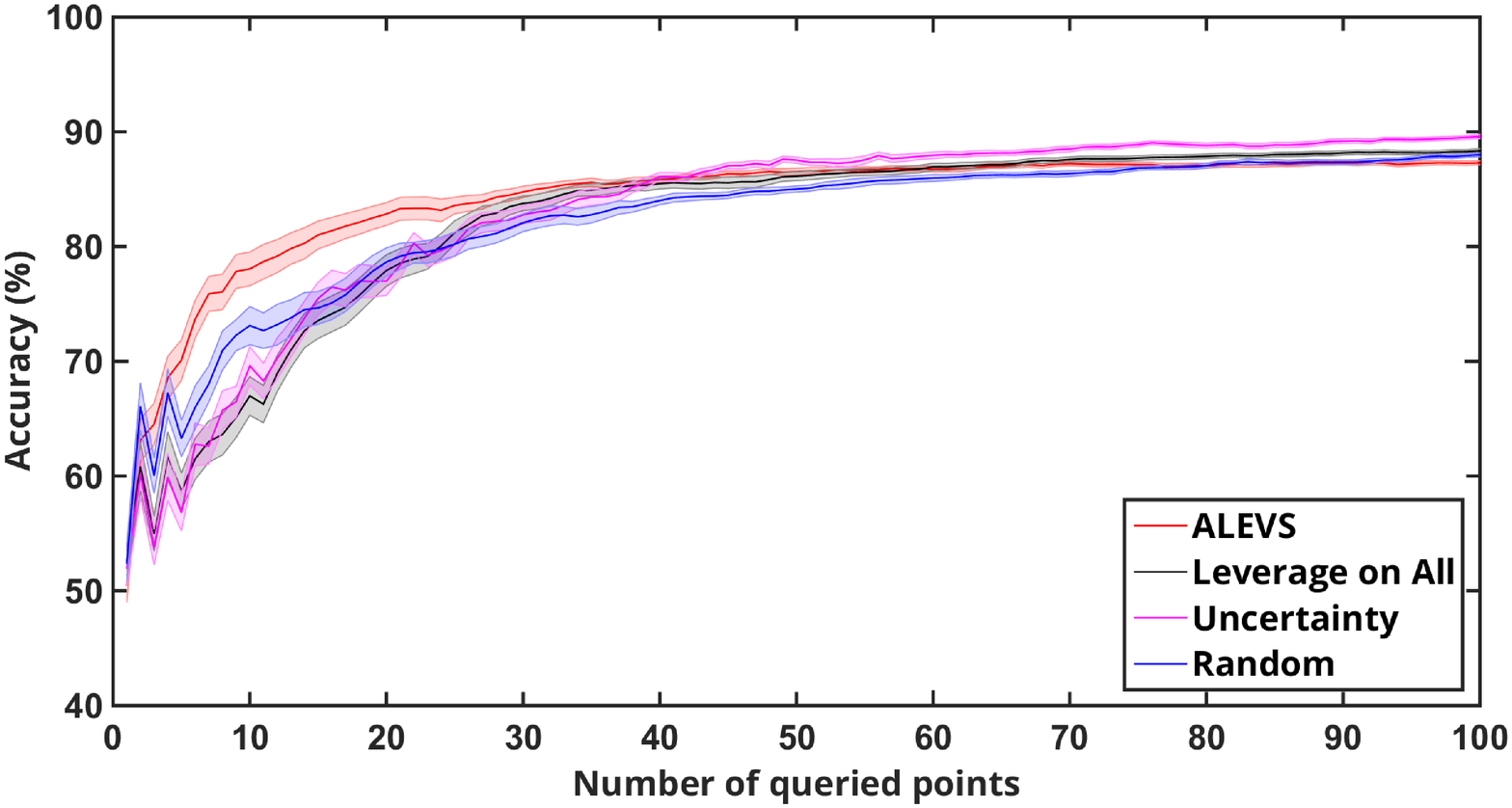}
		\caption{\textit{spambase}, $k=60$, RBF }
		\label{spambase}
	\end{subfigure}
	~
	\begin{subfigure}[b]{\newW\textwidth}
		\includegraphics[width=\textwidth]{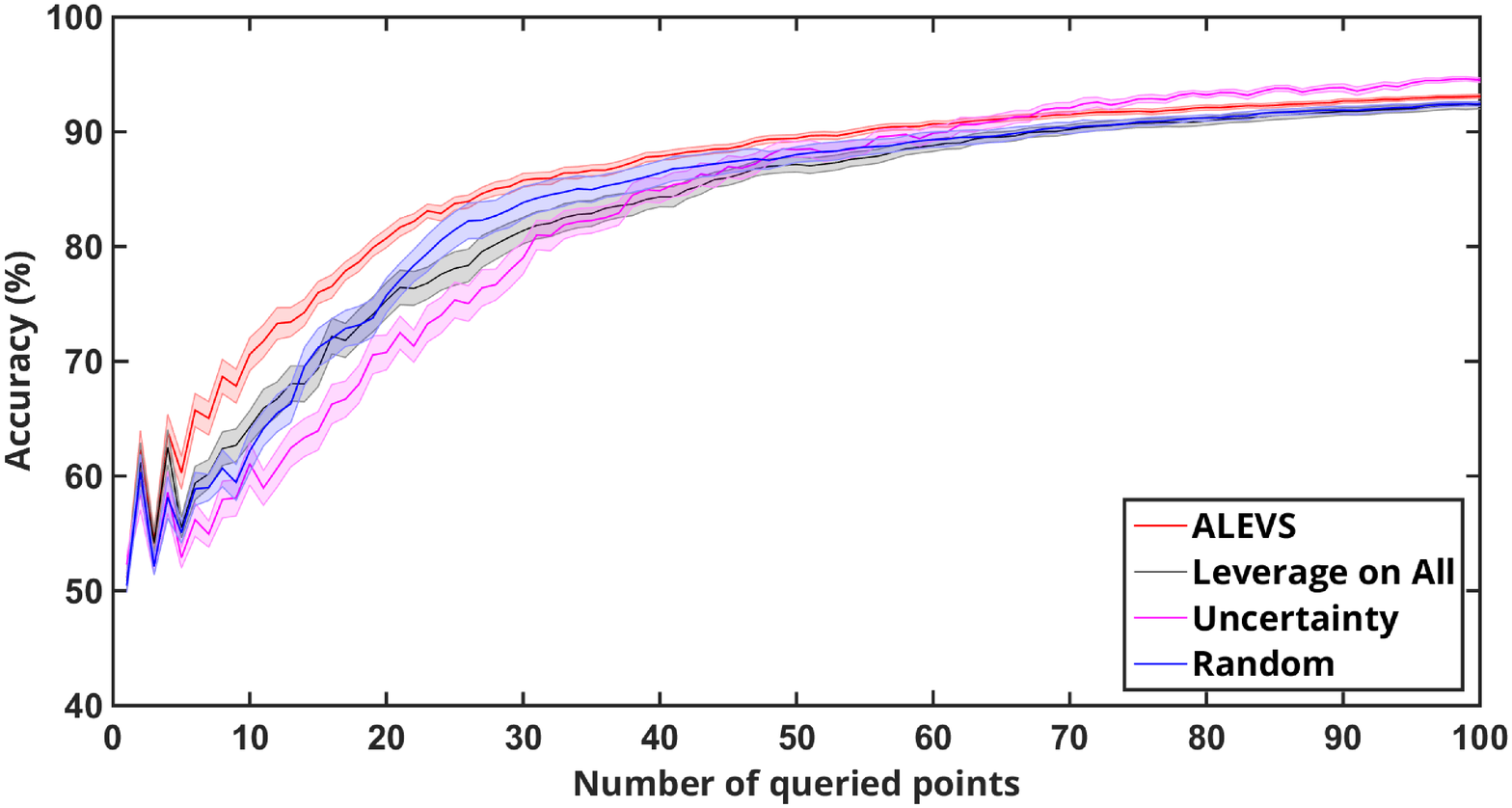}
		\caption{\textit{3 vs. 5}, $k=20$, linear}
		\label{MNIST}
	\end{subfigure}
	\caption{Comparison of ALEVS with baselines on classification accuracy.}
	\label{figs}
\end{figure*}

\section{Experiments}
\label{exps}

We compare ALEVS with the following baseline approaches: (1) Random Sampling: randomly select query instances, (2) Uncertainty Sampling: selects the instance with maximal uncertainty, (3) Leverage on all data: computes the leverage score on the $\mathcal{D}$ at the beginning of the iteration without paying attention to class membership and selects unlabeled queries in the order of their leverage scores. The last baseline decides whether separating the examples based on their predicted class membership has any value or not.

In Uncertainty Sampling, to find the most uncertain unlabeled datapoint based on the SVM output, we estimate the  posterior probabilities of each unlabeled instance with Platt's algorithm \cite{platt1999probabilistic}. The most uncertain point is the one with maximal $(1 - p( y^* \given{\vec{x}_i} ))$.

Ten different datasets are used in our study and their descriptions are given in Table 1.
The  \textit{digit1}, \textit{g241c}, \textit{g241n}, \textit{USPS} datasets are from \cite{chapelle2006semi}. The \textit{spambase} dataset and \textit{letter} are from \cite{Lichman:2013}. The \textit{letter} dataset is a multi-class dataset, we selected  letter pairs that are difficult to distinguish: \textit{letter(D vs. P)} and \textit{letter (U vs. V)}. Similarly, we work on \textit{MNIST(3 vs. 5)} which  is one of the most confused pairs in the handwritten digit dataset MNIST \cite{mnistlecun}. Finally, the \textit{splice} and \textit{ringnorm} are culled from  Gunnar R\"aetsch`s benchmark datasets \cite{ratsch2001soft}. In all experiments, an SVM classifier with RBF kernel is used as the classifier. For the RBF kernel scale parameter is selected automatically by a heuristic method of built-in SVM function in MATLAB.

\begin{table}[h]
\caption{Datasets and their dimensions.}
\vskip 0.15in
\begin{center}
\begin{small}
\begin{tabular}{|l|c|c|l|}
\hline
{\bf dataset}   & {\bf\# instances} & {\bf\# features}  \\ \hline
digit1      & 1500      & 241       \\ \hline
g241c        & 1500      & 241       \\ \hline
g241n         & 1500      & 241     \\ \hline
letter (DvsP)  & 1608      & 16       \\ \hline
letter (UvsV) & 1577      & 16        \\ \hline
USPS        & 1500      & 241    \\ \hline
splice         & 2991      & 60        \\ \hline
ringnorm      & 2000      & 20      \\ \hline
spambase      & 2000      & 57   \\ \hline
MNIST (3vs5)   & 2000      & 784    \\ \hline
\end{tabular}
\label{tbl:datasets}
\end{small}
\end{center}
\vskip -0.1in
\end{table}
Each dataset is divided into two portions at random. The first portion is held-out for testing purposes and the other half is used for training. We start with 4 initially labeled examples. At each iteration, the classifier is updated for all methods and the accuracies are calculated on the same held-out test data. For each dataset the experiment is repeated $50$ times and for each replicate, the partitioning of the whole data into training and test sets is random. The accuracies reported in figures are the average accuracies over these random trials with shaded area representing standard error. In calculating leverage scores we experimented with both RBF and linear kernel. Here we report the best performing cases.
\vspace{-3mm}
\section{Results}
% rewrite
Figure 1 shows the classification accuracy of ALEVS and the baselines with varied numbers of queries. We observe that in seven out of  ten datasets (Fig. a, b, c, g-j), ALEVS is able to outperform the baseline methods. In three datasets, the performance is comparable to that of uncertainty. In \textit{USPS} dataset (Fig. f), ALEVS beats Random Sampling and Leverage on All, however it is performance is only as good as Uncertainty Sampling. In \textit{letter(D vs. P)} (Fig. d ) and \textit{letter(U vs. V)} (Fig. e) dataset, the initial performance of ALEVS is very good, but as the number of queries increased Uncertainty Sampling outperforms ALEVS. In all results, at early iterations, ALEVS seems to query better data points. One strategy could be start with ALEVS and switch to another sampling strategy at further iterations.

The baseline Leverage on All sampling achieves a performance in between ALEVS and Uncertainty Sampling. This method calculates leverage scores for the kernel matrix computed over all data, whereas ALEVS first forms partitions based on the class membership. From the results, we conclude that this division is valuable. It might even be interesting to further divide data into clusters and calculate leverage scores of examples within their own clusters.

We probed the effect of $k$ parameter to the resulting performance. In the experiments, we operated with $k$ values 20, 40, 60 and 80. For the sake of simplicity for each dataset, we include results with best $k$ values. We observe that for \textit{USPS} and \textit{splice} datasets, $k$ affects the accuracy drastically. In our future line of work, we will investigate systematic means to set the parameter $k$ based on the input matrix structural properties. 

\section{Conclusion}
\label{conc}
In this paper, we propose a new method, \textsc{ALEVS}, that samples data points based on their statistical leverage scores. The leverage scores are calculated on kernel matrices constructed from the feature vectors of the instances. Empirical comparison with baseline methods demonstrates that sampling high-leverage points are indeed useful. In addition to the future work discussed in the Results section, we consider improving the computational efficiency. Since the input data matrices to the leverage score computation have overlap across iterations, we will investigate ways of reusing leverage computations in previous iterations to calculate the leverage scores for the current iteration.

\vspace{-3mm}
\section*{Acknowledgments} 
\vspace{-2mm}
{\small O.T. acknowledges support from Bilim Akademisi - The Science Academy, Turkey under the BAGEP program and the support from  L'Oreal-UNESCO under the UNESCO-L'OREAL National Fellowships Programme for Young Women in Life Sciences.}

\bibliography{ref}
\bibliographystyle{icml2015}

\end{document}